# Federated AI lets a team imagine together: Federated Learning of GANs

Rajagopal. A[1], Nirmala. V[2]

*Abstract—* Envisioning a new imaginative idea together is a popular human need. Imagining together as a team can often lead to breakthrough ideas, but the collaboration effort can also be challenging, especially when the team members are separated by time & space. What if there is a AI that can assist the team to collaboratively envision new ideas?. Is it possible to develop a working model of such an AI? This paper aims to design such an intelligence. This paper proposes an approach to design a creative & collaborative intelligence by employing a form of distributed machine learning approach called Federated Learning along with Generative Adversarial Network (GAN) fusion. This collaborative creative AI presents a new paradigm in AI, one that lets a team of two or more to come together to imagine and envision ideas and one that synergies well with each other's likes. This is possible by a new way to combine federated learning with a new way to combine multiple GANs together. In short, this paper explores the design of a novel type of AI, called **Federated AI Imagination,** one that lets geographically distributed teams to collaboratively imagine.

## I. INTRODUCTION: FEDERATED AI IMAGINATION

### A. Colloborative imagination by a geo distributed team

Table 1: How is this paradigm different?

|  | *Functionality* | *References* |
|---|---|---|
| **Current state of art (2017 to May 2019)** | AI learns using a set of users | FL by Google [1] |
| **This paper** | A set of users use AI to Imagine together | FL & GAN |

So far Federated Learning (FL) by Google[1] performs classification tasks by learning from a group of user's data. But what if the goal of AI is help a team of users to jointly imagine together? Is this possible for AI to do?.

The main contribution of this paper is to add a new flavor to Federated Learning (FL) research, as illustrated in Table 1. By exploring a new paradigm of an AI that enables collective human imagination, a new wave of possibilities are let open.

The 5 major contributions of this paper are
1. Give a new flavor to Federated Learning research, as illustrated in Figure 2
2. Expand the scope of what is achievable by AI as per Table 3
3. Proposed a novel Deep Learning architecture for Collaborative Imagination for a team.
4. Demonstrate the feasibility of the proposed architecture by developing a working prototype.
5. Showcase the possibilities of Federated AI Imagination.

Is it possible for a team to use AI to imagine ideas. This paper explore this new paradigm, and demonstrates it is feasible.

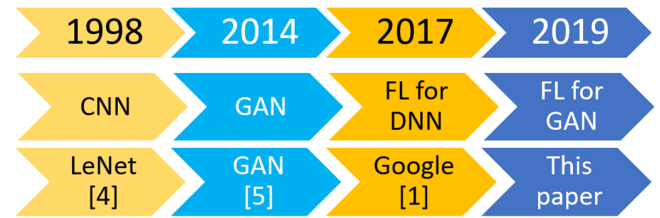

Figure 1. Trends in Deep Learning

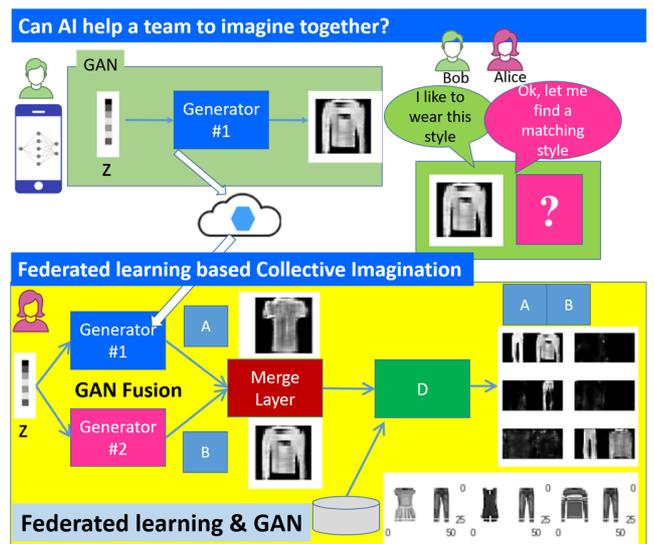

Figure 2: FL and GAN fusion, a new neural net for paired object synthesis

### B. Related work

Table 2: Federated Learning research

|  | Functionality | Deep Learning architecture | References |
|---|---|---|---|
| 2017 to May 2019 | A set of users learn together | TensorFlow Federated framework by Google [2] | FL by Google [1] |
| New Potential | A set of users Imagine together | Focus of this paper | This paper |

FL [1] introduced by Google in 2017 is an active research area. In 2019, Google released TensorFlow Federated at "tensorflow.org/federated". So far, FL has been used in classification tasks. But can FL be used in synthesis tasks? This paper explores this new possibility. How to design Deep Learning architecture for combined imagination by a team?

This paper combines Federated Learning with Generative Adversarial Networks (GAN) [3].

### C. Federated AI imagination enables new AI use cases

Table 3. New possibilities for AI

| What type of problems can be solved by the proposed AI approach? | | |
|---|---|---|
| **Federated AI Imagination** | | |
|  | *Proposed AI Capability* | *Example Challenges* |
| 1 | *Collaborative Planning (Collaborative Envisioning)* | A husband and wife want to plan which apparel to wear for the party, based on an understanding of the typical dress patterns of others. |
| 2 | *Paired Image synthesis* | Which couples usually go together? |
| 3 | *Sports partnership analysis* | In sports, which players partner together most? |
| 4 | *Collaborative Contemplation* | In smart city security, if you spot a terrorist in a camera in one part of the city, where is his partner? |
| 5 | *Imagining the future based on Cause and Effects.* | Imagining a Future event, given a particular event has happened |

This paper opens up the immense potential for AI in many dimensions.
1. AI for Collaborative Envisioning (Planning)
   - Figure 2 & 5 illustrates how a family thinks together to plan for an occasion
2. AI for Collaborative Deliberation (Contemplation)
   - Figure 3 illustrates how multiple city security cameras federate and imagine together to improve security of the city.
3. AI for Cause and Effect prediction and Visualization
   - Figure 8 explores how cause and effect relations across geo distributed events.

## II. FEDERATED AI IMAGINATION: USE CASE & DESIGN

What is the secret formula behind this AI paradigm? How to realize the possibilities of Table 3. A novel Deep Learning architecture to accomplish this is contributed by this paper. The idea behind this architecture is to combine the power of FL and Generative Adversarial Networks (GAN) multimodal fusion. This paper designs the deep learning architecture for the above AI paradigms and develops a working model. This novel neural network architecture is presented in Figure 3 and Figure 7.

### A. Federated AI Imagination Paradigm: Understanding it through a use case

The concept of **Federated AI Imagination** is illustrated in Figure 3 with a use case. How to design a city scale AI architecture that allows a collective team to perform a team activity. The challenge here is AI for counter-terrorism

### B. AI that Colloboratively Imagines

At the intersection of two promising active research areas, FL & GAN based Neural Network Fusion approaches, is the discovery of the promise of AI that powers **Collective Thinking by a team**. Architecture for combine GAN Fusion and FL is discussed in Figure 3. An implementation of GAN Fusion on is shown in Figure 5. As per Figure 5, key tricks are both Generators are sampled from the same noise vector, Generator #1 is frozen, so that Generator #2 learns during back propagation. Discriminator uses a new type of dataset, which has possible combinations of the 2 images. The working prototype is presented in Figure 4.

.

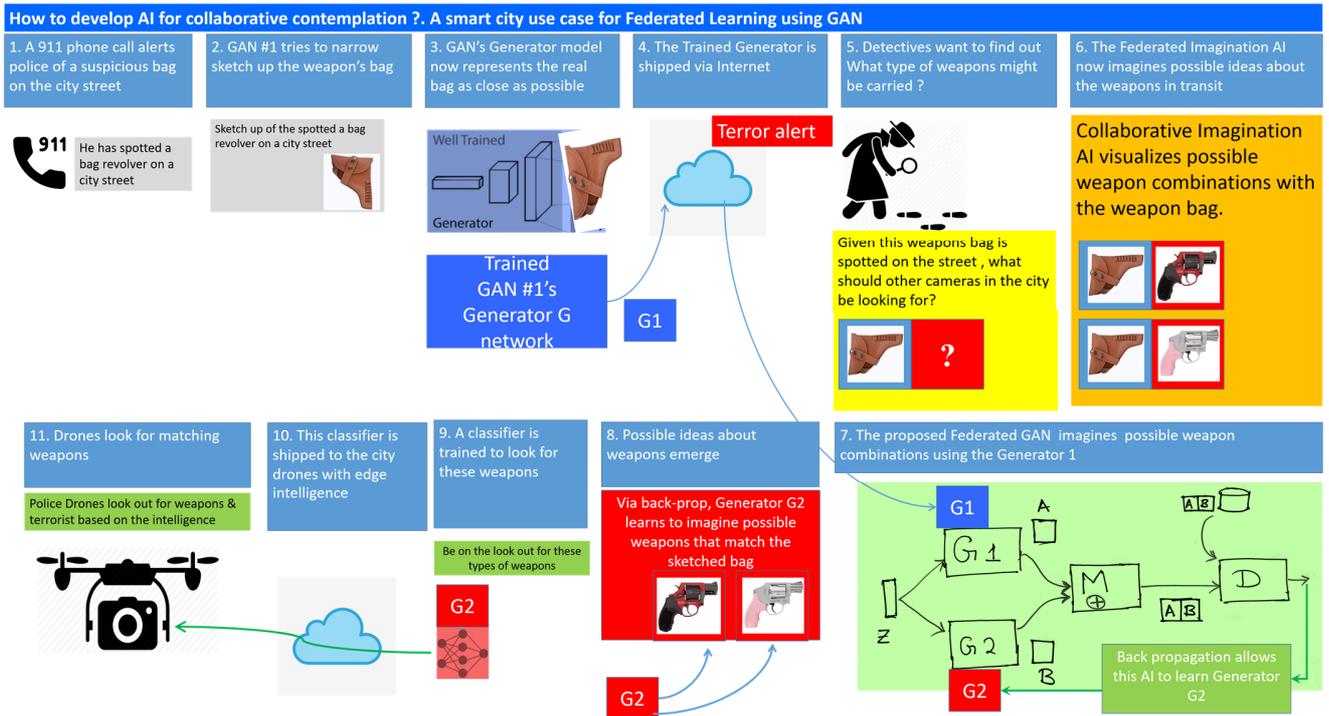

Figure 3. A use case to illustrate how to use AI for collaborative thinking.

*C. Algorithm for Federdated Imagination: A technique to combine FL & GAN Fusion*

1. A GAN based Generator G1 is trained to sketch up a weapon's visual that was spotted.
2. G1 is trained & uploaded into cloud to facilitate FL.
3. The model G1 get downloaded over internet and gets dynamically federated into a GAN Fusion neural net, as shown in module 4 in Figure 3.
4. The GAN Fusion model merges few generators, $G^i$ from multiple users. Each G can be a trained by a user on another device and dynamically federated over internet. The design for GAN Fusion is shown in module 7 in Figure 3. G1 has insights about imaginative sketch of bag as imagined by first user. G1 is frozen inside the Fusion network as they represent user's interest. Next, the goal is to use back propagation on the Fusion network. This allows G2 to learn, given G1 is non-trainable now.
5. After learning, G2 will be able to imagine possible weapons that would have been carried in that bag.
6. This intelligent imagination patterns by this Federated AI is transmitted to drones for city surveillance.

The architecture showed how imagination power of humans is handed over internet into a Fusion GAN, allowing further imagination. This is the essence of the idea of using FL with GAN Fusion. Thus Federated AI Imagination by a team has been designed.

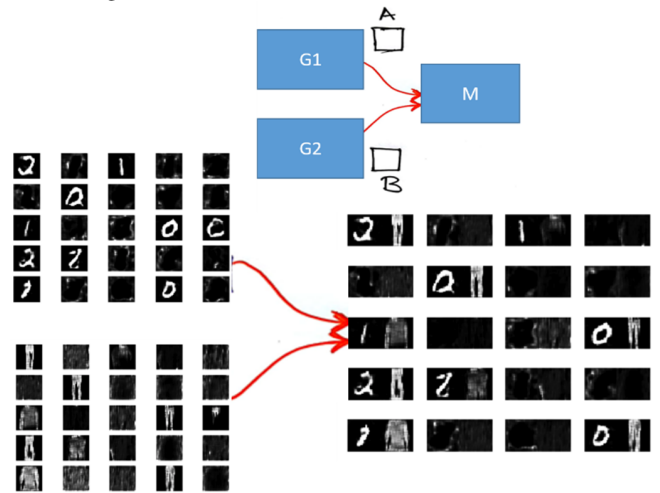

Figure 4. Screenshot of output of GAN Fusion.

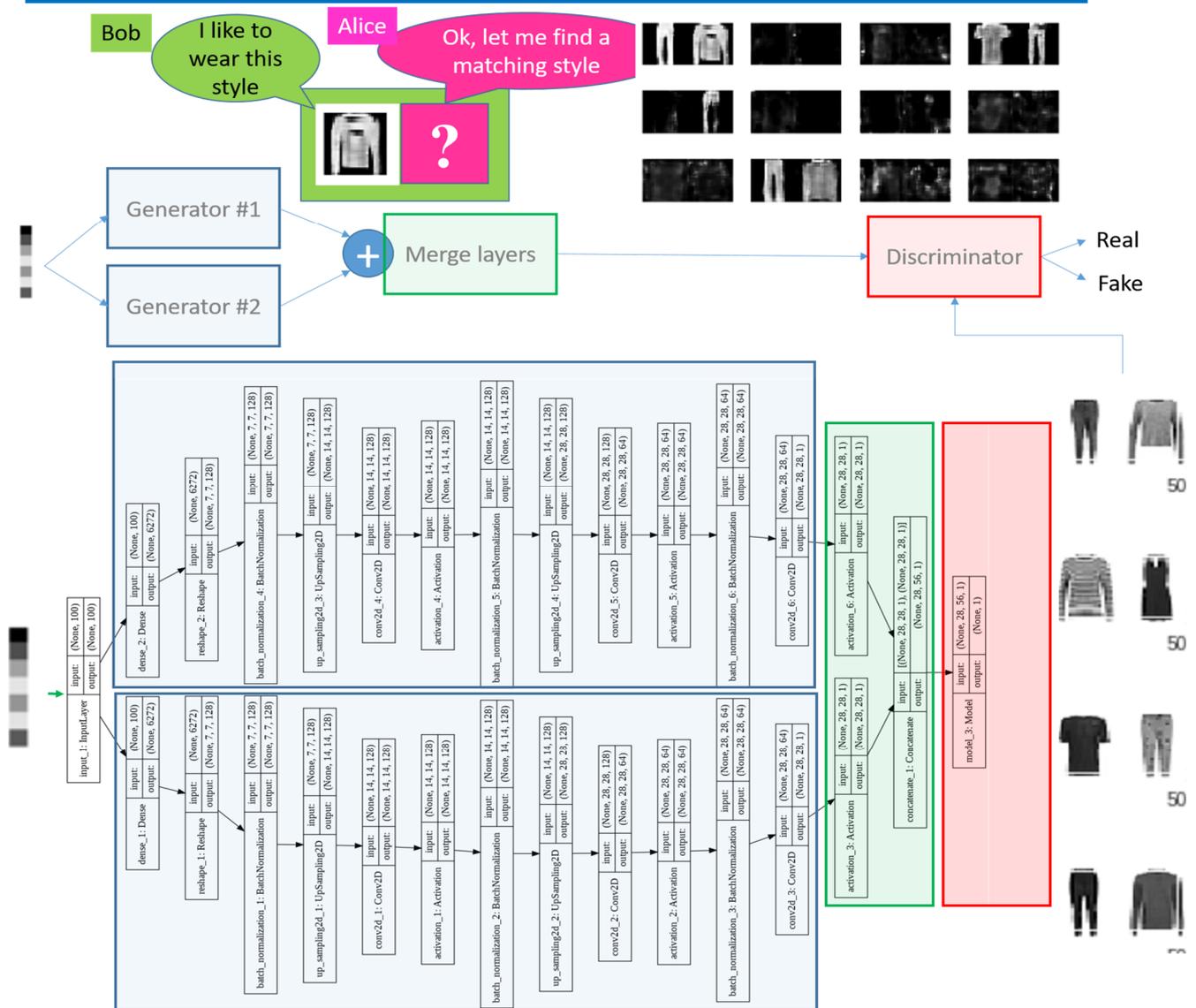

Figure 5. Implementation of GAN Fusion architecture. This powers Collaborative AI Imagining by a team

Table 7. Nomenclature used in this paper

| NOMENCLATURE | |
|---|---|
| GAN | Generative Adversarial Network [3] |
| FL | Federated Learning [1] |
| G | Generator of a GAN |
| G1 | Refers to model generator G of 1st user |
| G2 | Refers to model generator G of 2nd user |
| M | Merge layer (keras) |
| D | Discriminator of GAN |
| TFF | TensorFlow Federated Framework by Google |

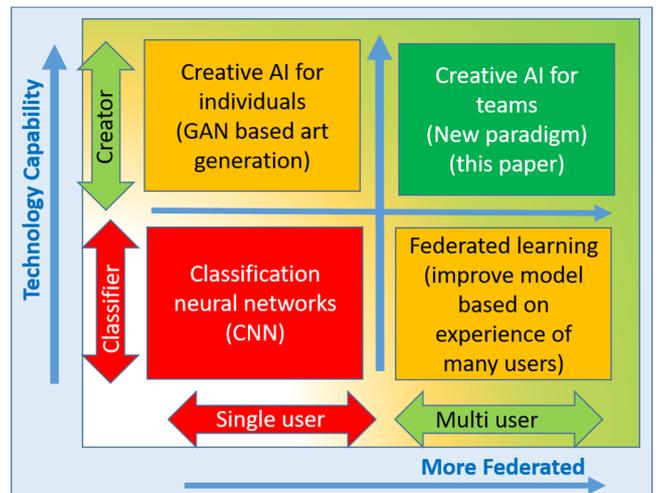

Figure 6. This work is at the juncture of 2 promising research areas, FL & GAN Fusion

## III. CAN AI UNDERSTAND & THINK OF DISTRIBUTED EVENTS ACROSS GEOS

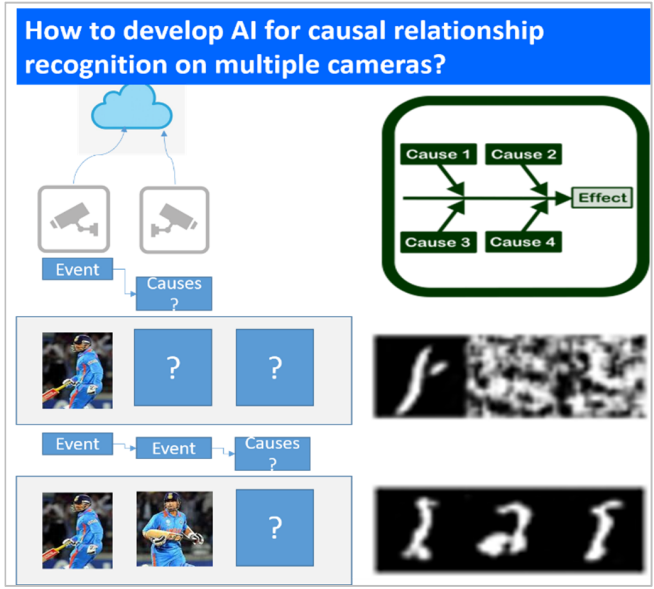

How to develop AI that predicts the future from a set of geo distributed robots/cameras? Is it possible for a GAN Fusion to find patterns of cause and effect? Is it possible to create an visualization of the future given past events across many parts of the city?. The challenge presented in Figure 7 is approached by the formula of innovatively repeating a geo distributed computation graph. This algorithm is listed in Table 4. By repeating the FL based GAN Fusion distributed computation graph at periodic intervals, this AI achieves the feat of picturing the future based on patterns observed in the past across different parts of the city.

Table 4. FL based Deep Learning algorithm for predicting the future based on events happening across spatial dimensions

| | | Federated Deep Learning Algorithm to correlate events across geos and visualize the future |
|---|---|---|
| 1 | | 2 stages of GAN Fusion is hired, as shown in Figure 8 |
| | 1.1 | In the 1st stage of GAN Fusion, G2 learns to visualize the future based on intelligence gather by G1 |
| | 1.2 | Once the 1st stage is trained, G1 and G2 are transmitted over the internet for fusion at next stage |
| | 1.3 | In the 2nd stage of GAN Fusion, G3 learns to visualize the future based on happenings in previous 2 locations as represented by G1 & G2 |
| 2 | | Thus, Generator G3 learns the causal relationship between 3 locations |
| 3 | | G2 even can start predicting based on trick listed below |
| 4 | | G3 even can start predicting and visualizing the future by the below trick |
| | 4.1 | Every few minutes, the Federated learning transmits G1 and G2 |
| | 4.2 | Meanwhile G1 & G2 counties to learn |
| | 4.3 | Then G3 learns to match patterns in the 3 locations based on intelligence of G1 & G2 |
| 5 | | Repeat steps 1 to 4 again in a loop |
| 6 | | G uses a spatio temporal model, I3D / 3DCNN [15] |

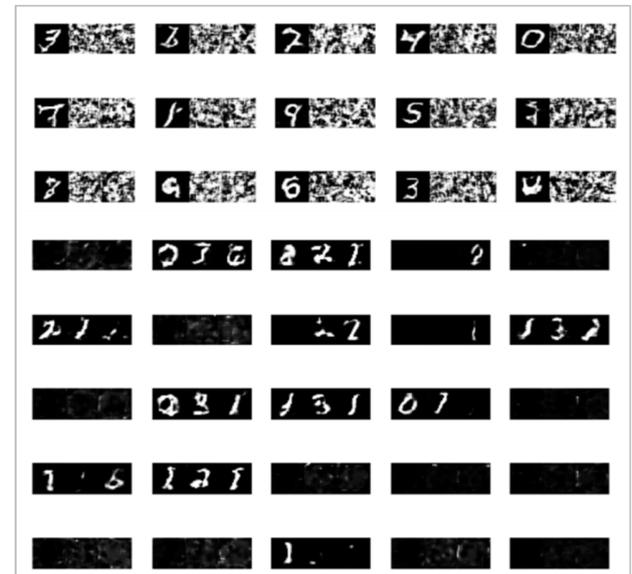

Figure 7. Screenshot of output for prototype of an AI that visualizes future events by combing intelligence gathered across multiple locations.

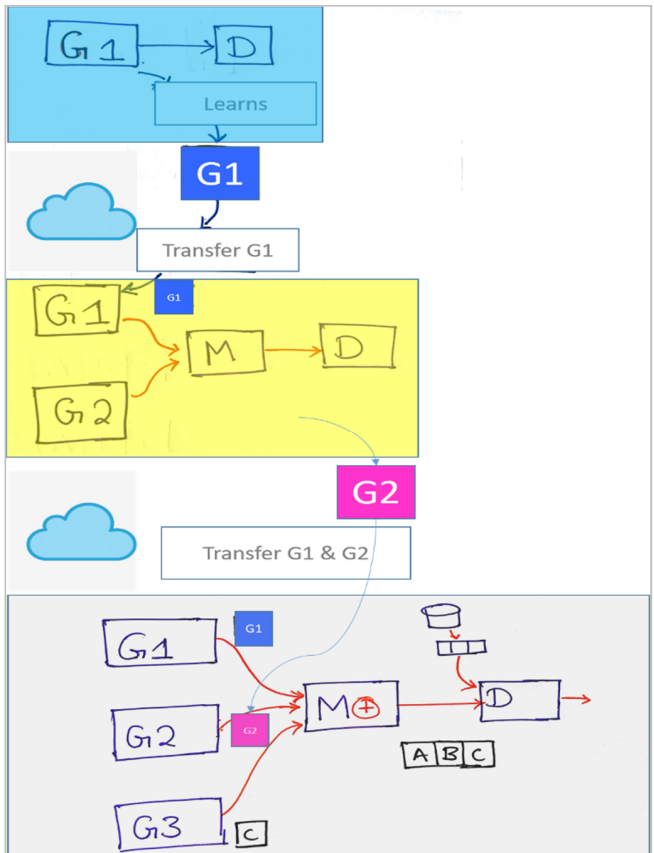

Figure 8. Federated Learning over cloud enables multiple users to participate during Collaborative Consulting.

## IV. RESULTS AND CONTRIBUTIONS

The potential for Federated AI Imagination was explored in this work.

Key results are
1. Contributed a new flavor to FL research as per Table 5
2. Expanded the potential of Federated AI as per Table 6
3. Contributed a novel Federated Deep Learning approach for Collaborative Imagination for a team as per Figure 7.
4. Demonstrated feasibility of Federated Imagination with a working prototype. Figure 5 shows screenshots.

Table 5. Key contribution in Distributed Machine Learning research

| Paradigm | Objective of AI | Design | Benefits |
|---|---|---|---|
| *Federated Learning* | Learn to pattern by training on multiple user's data for better recognition classification | FL with DNN or CNN | AI learns while protecting user's Privacy |
| *Proposed architecture* | Collaborative Imagination / Synthesis / | FL with GAN | Collaborative planning tasks |

## V. CONCLUSION AND FUTURE SCOPE

Table 6. New uses cases unlocked

| | Expanding the possibilities of Federated AI | |
|---|---|---|
| | **Contributed AI Capability** | **Results** |
| 1 | *Collaborative Planning* | AI Capability is demonstrated as per Figure 5. |
| 2 | *Collaborative Contemplation* | AI Capability is demonstrated as per Figure 3. |
| 3 | *Imagining the future based on Cause and Effects.* | AI Capability is demonstrated as per Figure 8. |

Much promise holds for the future of AI at the intersection of FL research and GAN research. Federated Deep Learning with GAN Fusion is a new paradigm and sets the stage for new dimension in AI research, as discussed in Table 5 and 6.

## REFERENCES

[1] Smith, Virginia, Chao-Kai C, Maziar S, and Ameet S, *"Federated multi-task learning"*, Advances in Neural Information Processing Systems, pp. **4424**-**4434**. **2017**.

[2] Bonawitz et al., *"Towards Federated Learning at Scale: System Design."*, arXiv:**1902.01046**, **2019**

[3] Goodfellow Ian et al., *"Generative adversarial nets."*, Advances in neural information processing systems, pp. **2672-2680**, **2014**

[4] Kurakin, Alexey, I. Goodfellow, S. Bengio. *"Adversarial machine learning at scale."* arXiv:**1611.01236**, **2016**

[5] Konecny J, H. Brendan, Daniel R, and Peter R, *"Federated optimization: Distributed machine learning for on-device intelligence"*, arXiv:**1610.02527**, **2016**.

[6] Hard at el., *"Federated learning for mobile keyboard prediction"*, arXiv:**1811.03604** , **2018**.

[7] Zhang, H., Goodfellow, I., Metaxas, D. and Odena, A., *"Self-attention generative adversarial networks"*. arXiv:**1805.08318**, **2018**

[8] Wang, Xiaofei, Yiwen H, Chenyang W, Qiyang Z, Xu C, Min C, *"In-edge AI: Intelligentizing mobile edge computing, caching and communication by federated learning."*, arXiv:**1809.07857**, **2018**

[9] Ma, Jiayi, Wei Yu, Pengwei Liang, Chang Li, and Junjun Jiang, *"FusionGAN: A generative adversarial network for infrared and visible image fusion."* Information Fusion **48**, pp. **11-26**, **2019**

[10] Bonawitz at el., *"Practical secure aggregation for privacy-preserving machine learning."* Proceedings of the 2017 ACM SIGSAC Conference on Computer and Communications Security, pp. **1175-1191**, ACM, **2017**

[11] Andrew Y. Ng et al, *"Multimodal deep learning."*, Proceedings of the 28th international conference on machine learning, ICML-11, pp. **689-696**, **2011**

[12] Zhang at el, *"Stackgan: Text to photo-realistic image synthesis with stacked generative adversarial networks."*. Proceedings of the IEEE International Conference on Computer Vision, pp. **5907**-**5915**, **2017**

[13] Jarrahi, M. Hossein. *"Artificial intelligence and the future of work: human-AI symbiosis in organizational decision making."*, Business Horizons 61, no. 4 , pp **577**-**586**, **2018**

[14] Nie et al., *"Medical image synthesis with context-aware generative adversarial networks."*, International Conference on Medical Image Computing and Computer-Assisted Intervention, pp. **417**-**425**. Springer, Cham, **2017**

[15] N.S. Lele *"Image Classification Using Convolutional Neural Network"*, International Journal of Scientific Research in Computer Science and Engineering **6**.**3**, Pp. **22**-**26**, **2018**

[16] Bala , Sudhakar, *"Privacy-Preserving Data Transmission protocol for Wireless Medical Sensor Data"*, International Journal of Scientific Research in Computer Science and Engineering, Vol.**5**, Issue.**3**, pp.**132**-**135**, **2017**

[17] Mishra et al, *"A generative approach to zero-shot and few-shot action recognition."*, IEEE Winter Conference on Applications of Computer Vision (WACV), pp. **372**-**380**. IEEE, **2018**

**Comments**

1 rajagopal.motivate@gmail.com
Indian Institute of Technology, Madras
2 gvan.nirmala@gmail.com (correspondance)
Queen Mary's College